\documentclass{article}
\usepackage{ijcai25}

\usepackage{times}
\usepackage{soul}
\usepackage{url}
\usepackage[hidelinks]{hyperref}
\usepackage[utf8]{inputenc}
\usepackage[small]{caption}
\usepackage{graphicx}
\usepackage{amsmath}
\usepackage{amsthm}
\usepackage{booktabs}
\usepackage{algorithm}
\usepackage{algorithmic}
\usepackage[switch]{lineno}

\newtheorem{example}{Example}
\newtheorem{theorem}{Theorem}

\title{Dynamic Replanning for Improved Public Transport Routing\thanks{This is a preprint of a paper accepted for publication at IJCAI 2025. The final version will appear in the IJCAI 2025 proceedings.}}

\author{
Abdallah Abuaisha\and
Bojie Shen\and
Daniel Harabor\and
Peter Stuckey\And
Mark Wallace\\
\affiliations
Department of Data Science and AI, Monash University, Australia\\
\emails
\{abdallah.abuaisha, bojie.shen1, daniel.harabor, peter.stuckey, mark.wallace\}@monash.edu
}

\usepackage{color}
\usepackage{xcolor}
\usepackage{amsthm}
\newtheorem{defn}{Definition}

\newcommand{\ignore}[1]{}
\usepackage{soul} 
\usepackage{booktabs} 
\usepackage{multirow}

\usepackage[algo2e,ruled,noend,linesnumbered]{algorithm2e}

\usepackage{etoolbox}
\makeatletter
\patchcmd{\@algocf@start}
  {-1.5em}
  {0pt}
  {}{}
\makeatother

\begin{document}

\maketitle

\begin{center}
    \textbf{Preprint.} Accepted to IJCAI 2025.
\end{center}

\vspace{10pt}

\begin{abstract}

    Delays in public transport are common, often impacting users through prolonged travel times and missed transfers. 
    Existing solutions for handling delays remain limited;
    backup plans based on historical data miss opportunities for earlier arrivals, while snapshot planning accounts for current delays but not future ones.
    With the growing availability of live delay data, users can adjust their journeys in real-time.
    However, the literature lacks a framework that fully exploits this advantage for system-scale dynamic replanning. 
    To address this, we formalise the dynamic replanning problem in public transport routing and propose two solutions: a ``pull" approach, where users manually request replanning, and a novel ``push" approach, where the server proactively monitors and adjusts journeys.
    Our experiments show that the push approach outperforms the pull approach, achieving significant speedups.
    The results also reveal substantial arrival time savings enabled by dynamic replanning.

\end{abstract}


\section{Introduction}

Public transport is a crucial component of smoothly functioning urban mobility systems. 
Unfortunately, in dynamic environments, travel conditions can change rapidly and unexpected incidents can occur, making delays in public transport fairly common. 
These delays often result in missed transfers, prolonged travel times, and late arrivals for users.
Efficient routing that accounts for real-time delays can play a vital role in reducing travel times for users, ultimately enhancing their satisfaction and trust in the public transport system. 

To handle dynamic scenarios, many works utilise historical delay data or delay probability distributions.
Backup plans (i.e., a policy) are precomputed offline, with the objective of maximising reliability by providing plans that remain viable when potential delays occur, or minimising expected arrival time at the destination~\cite{botea2013multi,dibbelt2014delay,redmond2022reliability}.
These plans, which can be printed in advance, offer multiple options at each transfer stop for users to consult in case of disruptions.
However, these contingent planning approaches fail to account for all possible scenarios and only address delays of limited duration, potentially missing opportunities for earlier arrivals.
Another family of algorithms demonstrates the feasibility of efficiently updating timetables \cite{cionini2017engineering,dibbelt2018connection} or precomputed data structure \cite{d2019dynamic,baum2023ultra} to account for delays. 
While this can enable snapshot planning with updated information, resulting journeys may be infeasible or suboptimal due to unforeseen future delays.

\begin{figure} [t]
    \centering
    \includegraphics[width=0.85\columnwidth]{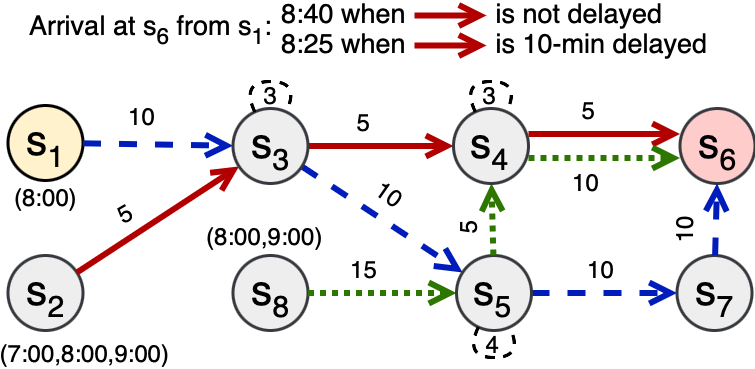}
    \caption{ 
        A toy network with eight stops ($s_1$ to $s_8$), where the origin $s_1$ is yellow and the destination $s_6$ is pink.
        Routes $r_1$, $r_2$, and $r_3$ use dashed blue, solid red, and dotted green arrows, labelled with travel times in minutes.
        Scheduled departure times for trips on each route are noted at its starting stop.
        Loops at stops indicate transfer times.}
    \label{fig:example}
\end{figure}

With widespread internet access, many transport operators and journey-planning applications, such as Google Maps, provide real-time data, including service disruptions and updated vehicle departure and arrival times. 
This enables users to manually review updates and request replanning when necessary.
Dynamic replanning, a widely adopted concept, enables real-time adaptation to environmental changes, addressing the limitations of offline and snapshot planning.
By leveraging updated information and computational techniques, it enhances efficiency and reliability in applications such as robotics ~\cite{simmons1992concurrent} and multi-agent systems ~\cite{YueExcution}.
However, extending dynamic replanning to public transport routing is challenging.
The number of replans and the search space required for a single journey often exceed expectations. 
Not only disruptions directly impacting the journey (e.g., a missed transfer or a delayed service) must be considered, but also those affecting other services in the network, which can sometimes lead to an earlier arrival time at the destination.
Consequently, frequent replanning is essential, with each replan accounting for all timetable services.

\begin{example}\label{ex:motiv}
Consider Figure \ref{fig:example}. 
Departing from the origin stop $s_1$ at 8:00, the original optimal plan, according to the normal timetable, involves taking the 8:00 trip on route $r_1$, $t_1$, and arriving at the destination stop $s_6$ at 8:40.
However, if the 8:00 trip on route $r_2$, $t_2$, is delayed by 10 minutes while $t_1$ runs on schedule, then catching the delayed $t_2$ at stop $s_3$ becomes possible.
The updated optimal plan involves taking $t_1$ from $s_1$, then changing at $s_3$ to $t_2$, which arrives at $s_6$ at 8:25, earlier than the original plan's arrival time of 8:40.
\end{example}

To the best of our knowledge, we are the first to extend the dynamic replanning concept to public transport routing.
We begin by providing a comprehensive description and formulation of the problem. 
Next, we propose a straightforward \emph{pull} approach, where users frequently request real-time journey replans from a central server via their edge devices (e.g., mobile phones), utilising the typical centralised setup for managing queries. 
However, the pull approach has several drawbacks:
(i) it is inconvenient for users due to frequent manual actions, such as monitoring updates and requesting replans;
(ii) it relies on user judgement, which may lead to missed opportunities for earlier arrivals; and
(iii) it places pressure on the server with numerous requests, reducing efficiency and complicating timetable updates.
To address these issues, we further propose a novel \emph{push} approach, where the server continually monitors and adjusts user journeys, pushing revised plans to users' edge devices. 
While more convenient for users, simply replanning strains the server, as it must monitor and replan for all users based on the entire timetable.
To improve efficiency, the server creates a query-specific \emph{envelope}, containing only relevant timetable parts.
This envelope is sent to the edge device, which subscribes to envelope updates and performs local replanning on the envelope, reducing the search space and distributing the computational load.

Our extensive experiments across metropolitan datasets highlight two key findings:
(i) the push approach achieves significant query efficiency, delivering runtimes within a fraction of a second, an order of magnitude faster than the pull approach, while enabling the capacity to process 10–20 times more queries;  
(ii) our dynamic replanning strategy saves 5–30 minutes per journey on average compared to static and other dynamic strategies, even when the initial plan is not delayed.


\section{Preliminaries}
\label{sec:Pre}

\subsection{Timetable Modelling} 
The timetable is the main input for any public transport routing system. 
We follow one of the most popular methods, as detailed in ~\citeauthor{dibbelt2018connection} (\citeyear{dibbelt2018connection}), which directly utilises the timetable structure rather than constructing a graph.
Specifically, we represent a timetable as $TB = (S, T, C, F)$, where $S$ are Stops, $T$ are Trips, $C$ are Connections, and $F$ are Footpaths. Each component is defined as follows:
\begin{itemize}
    \item 
    A \emph{stop} $s \in S$ is a departure and/or arrival point where a vehicle stops to pick up and/or drop off passengers.

    \item  
    A \emph{trip} $t \in T$ corresponds to a scheduled transport vehicle that visits a specific order of stops.
    Trips with the same order of stops can be grouped into a \emph{route}.

    \item 
    Each trip $t \in T$ is divided into a sequence of \emph{connections}, denoted as $t^C$ = $\langle c_0, c_1, \dots, c_k \rangle$.
    A connection $c \in t^C$ is represented as a 5-tuple $(s_{dep}, \tau_{dep}, s_{arr}, \tau_{arr}, t)$, indicating an event where trip $t$ departs from stop $s_{dep} \in S$ at time $\tau_{dep}$ and arrives at stop $s_{arr} \in S$ at time $\tau_{arr}$, with no intermediate stops.
    Note that $s_{dep} \neq s_{arr}$ and $\tau_{dep} < \tau_{arr}$ always hold. 
    The connections in $t^C$ satisfy the following properties\footnote{We often use the notation $a(b)$ to denote ``a of b'', unless stated otherwise. For instance, $t(c_i)$ refers to the trip of connection $c_i$.}:
    (i) $t(c_0) = t(c_1) = \dots = t(c_k)$, (ii) $s_{dep}(c_i) = s_{arr}(c_{i-1})$, and (iii) $\tau_{dep}(c_i) \geq \tau_{arr}(c_{i-1})$, $\forall i \in \{1 \dots k\}$.
    The union of all connections from all trips in $T$ forms the full set of connections, $C$.

    \item 
    A \emph{footpath} $f \in F$ models walking between two stops to change trips.
    It is represented as a triple $f = (s_i, s_j, \Delta\tau(s_i,s_j))$, indicating a transfer from stop $s_i \in S$ to stop $s_j \in S$ with a duration of $\Delta\tau(s_i,s_j)$.
    We require each stop $s_i \in S$ to have a loop footpath $f = (s_i, s_i, \Delta\tau(s_i,s_i))$.
    We associate with each stop $s_i \in S$ a list of its outgoing footpaths as $f(s_i) = \{f_0, \dots, f_m\} \subseteq F$.
    Following~\citeauthor{dibbelt2018connection} (\citeyear{dibbelt2018connection}), we assume that footpaths naturally satisfy the transitive closure and triangle inequality properties.

\end{itemize}

Given timetable $TB$, a journey from origin stop $s_o$ to destination stop $s_d$, consists of a sequence of connections $j = \langle c_0, \dots, c_n \rangle$, where $s_{dep}(c_0) = s_o$ and $s_{arr}(c_n) = s_d$.
Each pair of consecutive connections either shares the same trip $t \in T$ (i.e., no vehicle change) or involves a transfer via a footpath $f \in F$.
Each transfer must respect the required transfer time.
Formally, $\tau_{arr}(c_{i-1}) + \Delta\tau(s_{arr}(c_{i-1}), s_{dep}(c_i)) \leq \tau_{dep}(c_i)$ must hold $\forall i \in \{1\dots n\}$.

\paragraph{Earliest Arrival Time Problem (EATP).} 
Given a query $q = (s_o, s_d, \tau_q)$, EATP aims to find a journey $j$ that departs from origin $s_o$ no earlier than time $\tau_q$ and arrives at destination $s_d$ as early as possible, at time $\tau_d$, assuming no delays in timetable $TB$.
Such a journey is denoted as $j(s_o, s_d, \tau_q)$.

\subsection{Delay Modelling} 
Due to the dynamic nature of public transport networks, real-time disruptions, such as traffic congestion, technical issues, and adverse weather, often lead to delays.
These delays can significantly disrupt static travel plans, resulting in missed transfers and increased travel times.
To model delays, we assume the transport system maintains a centralised server. In addition to storing the original timetable $TB$, the server also keeps a list of delay events $EV$.

\begin{defn} 
    (Delay Event): A delay event $ev \in EV$ is represented as a tuple $(t, \tau_\delta, \delta)$, where $t \in T$ denotes the trip experiencing the delay, $\tau_\delta$ specifies the time when the delay occurs, and $\delta$ represents the duration of the delay. The time $\tau_\delta$ must fall within the trip's timeframe (i.e., $\tau_{dep}(c_0) \leq \tau_\delta \leq \tau_{arr}(c_k)$ for $t = \langle c_0, \dots, c_k \rangle$).
\end{defn}

When a delay event occurs, it impacts all subsequent connections in trip $t$ after time $\tau_\delta$.
To avoid synchronisation issues, we assume the real-time timetable $TB_{curr}$ at time $\tau_{curr}$ can be derived from the original timetable $TB$ and the delay events $EV$.
This requires identifying each delayed trip $t$ based on $EV$, scanning its affected connections $c$ in $TB$, and adjusting their departure and arrival times according to the trip's delay $\delta$ (i.e., $\tau_{dep}(c) = \tau_{dep}(c) + \delta$ and $\tau_{arr}(c) = \tau_{arr}(c) + \delta$).
Unlike previous works, we do not impose any restrictions on the maximum duration of delays.

\paragraph{Snapshot Earliest Arrival Time Problem (SEATP).} 
Given a query $q=(s_o,s_d,\tau_q,EV)$, SEATP plans a journey $j$ as in EATP, but while accounting for current delay events $EV$ in a snapshot of timetable $TB$, denoted as $TB_{curr}$.

\subsection{Snapshot Solver}

There are several algorithms for solving the SEATP, with the Connection Scan Algorithm (CSA) \cite{dibbelt2018connection} being one of the most efficient online approaches. Below, we provide a brief overview of the CSA.

Unlike Dijkstra's algorithm \cite{DBLP:journals/nm/Dijkstra59}, CSA does not operate on a graph and does not require a priority queue.
Instead, it assembles all connections from a timetable $TB$ into a single array $C$, sorted by (updated) departure time (i.e., from earliest to latest), and efficiently solves queries by scanning through this sorted array. 
Specifically, CSA maintains an array $A$ to store the tentative earliest arrival time for each stop $s \in S$. 
Initially, $A$ is set to infinity for all stops. 
Next, $A$ at each stop $s_i$ that is walking-reachable from the origin stop $s_o$ via a footpath in $f(s_o)$ is updated to $\tau_q + \Delta\tau(s_o, s_i)$.
The algorithm then proceeds to scan the connections in $C$ in sequential order, starting from the first connection that departs at or after $\tau_q$.
For each scanned connection $c$, the algorithm first checks its reachability.
A connection is considered \emph{reachable} if there exists a way to catch it on time (e.g., $A[s_{dep}(c)] \leq \tau_{dep}(c)$).
If $c$ is reachable and can improve the arrival time at $s_{arr}(c)$, the algorithm updates $A$ at each accessible stop $s_j$ via an outgoing footpath from $s_{arr}(c)$, if $A[s_j]$ can be improved.
Recall that $s_{arr}(c)$ has a loop transfer, resulting in updating $A[s_{arr}(c)]$ too.
However, if $c$ is not reachable, the algorithm simply skips it and proceeds to the next connection.
This process continues until all connections in $C$ departing before $A[s_d]$ are scanned. Finally, the algorithm returns the arrival time at the destination stop, $A[s_d]$.
To extract the full journey, the algorithm is augmented with journey pointers or followed by a post-processing phase.
For further details, refer to~\citeauthor{dibbelt2018connection} (\citeyear{dibbelt2018connection}).


\section{Dynamic Replanning}
\label{sec:delays}

Both contingent and snapshot planning can miss opportunities for improving arrival times.
To address this, we propose a dynamic replanning solution.
Dynamic replanning concerns adapting and improving plans during execution to address uncertainties in constantly changing environments.
This approach has been successfully applied in fields such as robotics and autonomous vehicles, enabling effective responses to real-time changes.
Building on this success, we extend the concept to public transport routing. 
Users are initially provided with an optimal plan, which is proactively adjusted as necessary throughout their journey based on the latest delay information. 
The goal is to reduce user travel times and impact of disruptions.
A key observation is that users cannot take any action while the vehicle is in motion between stops.
Therefore, replanning just before arriving at each stop along the journey until reaching the destination is sufficient, providing the finest feasible time granularity.
In broad terms, our dynamic replanning strategy for solving a single query consists of the following steps:

\begin{enumerate}

    \item Update the original timetable $TB$ with the latest delay events $EV$ at the current time $\tau_{curr}$ to derive the real-time timetable $TB_{curr}$.

    \item Plan an optimal snapshot journey $j$ from the current stop $s_{curr}$ to the destination stop $s_d$ using the current timetable $TB_{curr}$ with a snapshot solver.

    \item Execute the first action of the planned journey $j$ by taking its first connection $c$.

\end{enumerate}

Steps 1 to 3 are repeated, starting with the origin stop $s_o$ as the current stop $s_{curr}$, until the user reaches the destination stop $s_d$.
Let $c'$ be the connection taken in the previous iteration.
The action to be executed in the next iteration, considering connection $c$, can fall into one of four cases\footnote{To focus on more critical aspects of our algorithms, we omit cases (i) and (iv) in later discussions (but not in experiments); however, these can be easily handled by setting $c'$ and $c$ to null, resp.}:
(i) the user catches the trip with connection $c$ directly from $s_0$ as $c'$ is null;
(ii) the user simply stays on the current trip if $c$ and $c'$ belong to the same trip;
(iii) the user must transfer at stop $s_{arr}(c')$ if $c$ and $c'$ belong to different trips; or
(iv) the user can reach $s_d$ directly by walking via a footpath if $c$ is null.
Cases (i), (ii), and (iv) are straightforward.
For case (iii), we consider walking from $s_{arr}(c')$ to $s_{dep}(c)$ and taking connection $c$ as a single action, without any intermediate replanning, assuming the user is committed to this course of action.

Note that for the snapshot solver, offline preprocessing-based algorithms are generally unsuitable for frequent replanning due to the high cost of repairing precomputed auxiliary data \cite{bast2010fast,delling2015public,baum2023ultra,abuaisha2024efficient}.
In contrast, online algorithms, such as CSA \cite{dibbelt2018connection} and RAPTOR \cite{delling2015round}, are better suited for these scenarios, as they enable timetable updates at a relatively low computational cost.
In this paper, we adopt CSA as the underlying solver for replanning; however, any solver designed for a similar timetable structure could also be used.


\section{Our Approaches}

We introduce two approaches for dynamic replanning: a baseline pull approach and a more efficient push approach.

\begin{figure} [t]
    \centering
    \includegraphics[width=\columnwidth]{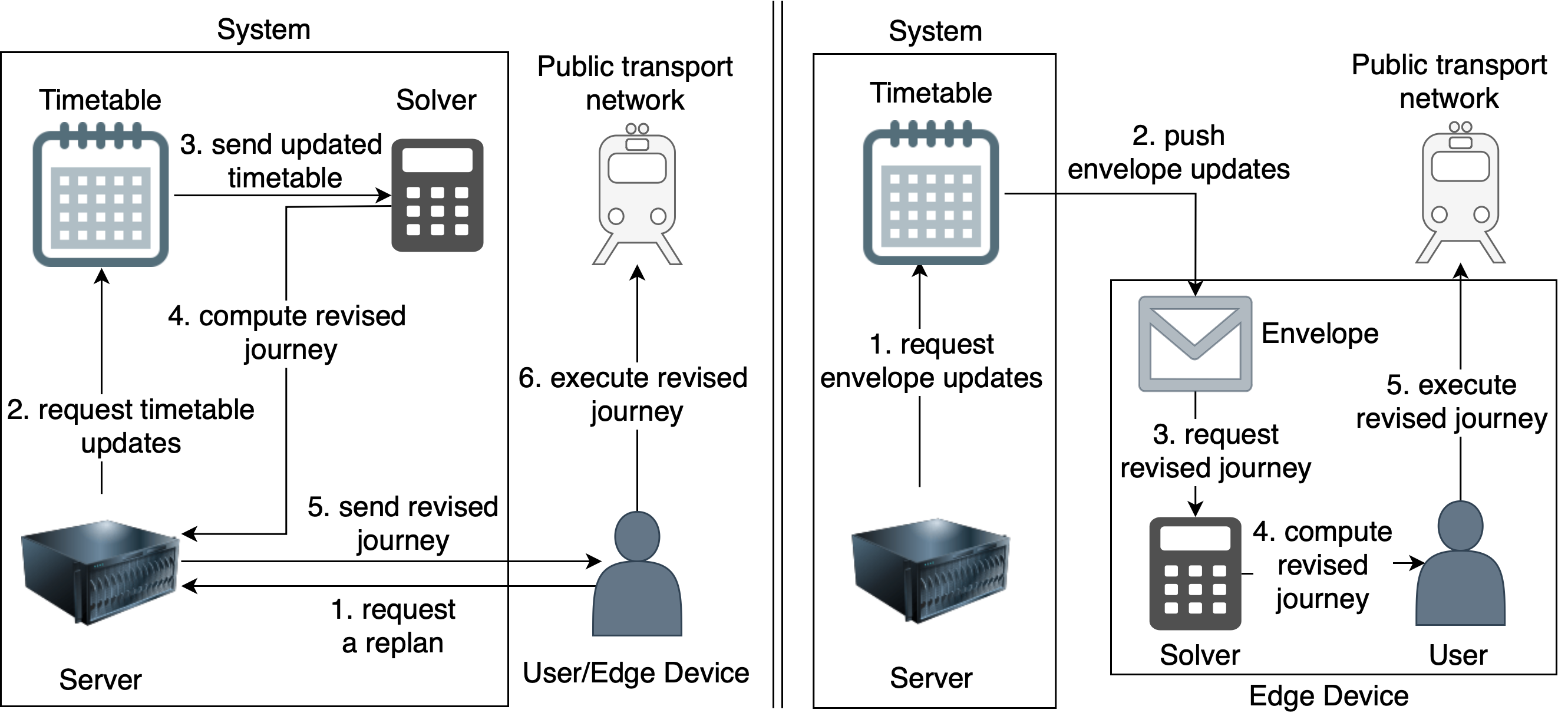}
    \caption{
        Framework for dynamic replanning during a typical replan step: pull approach (left) and push approach (right).
    }
    \label{fig:approaches}
\end{figure}

\subsection{The Pull Approach}

The straightforward approach follows the conventional setup where a central server handles all user queries in the network. 
In this approach, users frequently request (pull) journey replans from the server via their edge devices.
We assume users recognise the need to replan at every stop along their journey, in line with our formulation of dynamic replanning. 
To determine the next action the user should take, the server treats each request as a new query, computing an optimal journey from scratch based on the latest delay updates.
Figure \ref{fig:approaches} illustrates the framework of the pull approach for a single request.

For each request, the pull approach invokes the replanning algorithm to solve an SEATP query by setting $s_o$ to the user's current stop $s_{curr}$ and $\tau_q$ to the current time $\tau_{curr}$. 
Algorithm~\ref{algo:naive} presents the pseudocode for the pull approach's replanning algorithm, which returns the optimal journey toward $s_d$.
To begin, the algorithm retrieves the updated timetable $TB_{curr}$ based on the current time $\tau_{curr}$ (line~\ref{algo1:line:RetriveTimeTable}).
Next, as a prerequisite for running CSA, the algorithm sorts the connections $C$ in the timetable $TB_{curr}$ by their updated departure time, from earliest to latest (line~\ref{algo1:line:SortConnection}).
Once the sorting is complete, the algorithm executes CSA to plan an optimal journey $j$ from $s_{curr}$ to $s_d$, starting at the departure time $\tau_{curr}$ (line~\ref{algo1:line:RunCSA}).
Line~\ref{algo1:env} shown in grey is not used for the pull approach (see the next subsection).
Finally, the algorithm returns optimal journey $j$ (line~\ref{algo1:line:returnConnection}),
extended with the envelope when 
used in the next section.  
Note that if the first connection of the new journey $j$, $c_0$, does not belong to the same trip as the previous connection, a transfer from $s_{curr}$ to $s_{dep}(c_0)$ is included as part of the next execution.

\begin{algorithm2e}[t]
\small
\setcounter{AlgoLine}{0}
    \SetKwInput{Input}{Input}
    \SetKwInput{Output}{Output}
    \SetKwInput{Initialisation}{Initialisation}
    \Input{$s_{curr}$: current stop of user; $\tau_{curr}$: current time/departure time; $s_d$: destination stop;}
    \Output{$j$: optimal journey to $s_d$.}
    $TB_{curr} = $ retrieve real-time timetable at current time $\tau_{curr}$; \\ \label{algo1:line:RetriveTimeTable}
    Sort the connections in $TB_{curr}$ based on departure time; \\
    \label{algo1:line:SortConnection}
    $j = $ run CSA to plan an optimal journey $j(s_{curr}, s_d, \tau_{curr})$; \\
    \label{algo1:line:RunCSA}
    \color{gray} $Env =$ buildEnvelope$(s_{curr},s_d,\tau_{curr},\tau_d(j))$; \label{algo1:env} \\
    \color{black}
    \textbf{return} ($j$, \textcolor{gray}{$Env$});  \label{algo1:line:returnConnection}
    \label{algo1:return}
    \caption{Baseline Replanning Algorithm (Pull)}
    \label{algo:naive}
    \normalsize
\end{algorithm2e}

\subsection{The Push Approach}

Although the pull approach addresses dynamic replanning, it has significant drawbacks.
Users must repeatedly request a replan at each stop, which creates a cumbersome experience.
Additionally, the total number of requests can grow significantly, with each requiring a full timetable update and replan, which increases server load, especially for large timetables.

To overcome these challenges, we propose a novel approach in which the server continually monitors and adjusts user journeys, then sends (pushes) the revised plans to users' edge devices. 
To avoid replanning a user's journey from scratch using the entire timetable at every step, the server constructs an \emph{envelope} for each user.
This envelope contains only the relevant parts of the timetable that could affect the user's journey, limiting the search space for subsequent replanning steps.
Instead of storing a separate envelope for each user, which is resource-intensive, the server computes the envelope at the start of the query and transmits it to the user's edge device.
The edge device subscribes to the server for real-time envelope delay updates and performs local replanning as needed using the updated envelope.
This decentralised design distributes the workload between the server and edge devices, optimising resource utilisation and significantly reducing number of requests handled by the server.
Figure \ref{fig:approaches} illustrates the framework of the push approach for a typical replan step; note that the server may occasionally need to push a new envelope to the user, as explained later.

\subsubsection{Building the Envelope} 

The push approach recomputes the optimal journey at each replanning step until the destination is reached.
During each replanning, the optimal journey may change due to delays in only two scenarios: 
(i) the current optimal journey is delayed, causing a previously suboptimal journey to become the new optimal one by arriving earlier than the delayed journey; or
(ii) a previously unreachable connection is delayed, making it possible to catch it and form a new journey that arrives earlier than the current journey. 
In scenario (ii), the destination's arrival time of the current journey, $\tau_d$, serves as an upper bound.
The CSA only needs to scan through connections that could potentially form a journey which arrives earlier than $\tau_d$ to identify the new optimal journey. 
Based on this observation, we construct an envelope at the start of query $q$, that contains these connections to restrict the search space of CSA during replanning.
The envelope may need to be reconstructed later to accommodate additional delays affecting the current journey, as in scenario (i).
Fortunately, envelope reconstruction is infrequent, as shown in our experiments. We begin exploring envelope connections by constructing a time-independent graph, as defined below.

\begin{defn}
    (Time-Independent Graph (TIG)): 
    Given a timetable $TB = (S, T, C, F)$, the time-independent graph is a weighted directed graph $G = (V, E)$, where each vertex $v \in V$ represents a stop $s \in S$, and each edge $e \in E$ connects a pair of stops with the minimum duration among all connections and footpaths between them. Formally, $w(s_i,s_j) = \min (\{ \tau_j - \tau_i ~|~ (s_i, \tau_i,  s_j, \tau_j, \_) \in C\}
    \cup \{ \Delta\tau(s_i,s_j) ~|~ (s_i, s_j, \Delta\tau(s_i,s_j) \in F \})$. 
\end{defn}

\begin{example}
    The TIG for the network in Figure~\ref{fig:example} records the minimum cost for each edge. 
    It consists of edges $(s_1,s_3)$, $(s_3,s_5)$, $(s_5,s_7)$, and $(s_7,s_6)$, all with a duration of 10; 
    $(s_2,s_3)$, $(s_3,s_4)$, $(s_4,s_6)$, and $(s_5,s_4)$, all with a duration of 5; and 
    $(s_8,s_5)$ with a duration of 15.   
\end{example}

Based on the original timetable $TB$, 
the time-independent graph $TIG$ is precomputed during offline preprocessing and reused for all online queries.
A journey from $s_o$ to $s_d$ in $TB$, denoted as $j(s_o,s_d)$, can be represented as a path $p_G(s_o, s_d)$ in $TIG$.
Since $TIG$ assigns the minimum travel time of all connections and footpaths as the edge weight, the shortest path duration $sp_G(s_o, s_d)$ clearly serves as a lower bound for any journey from $s_o$ to $s_d$ in the timetable $TB$.
Given a query $q$ and the arrival time $\tau_d$ of the optimal journey at the destination, we 
construct the envelope as follows.

\begin{defn}(Envelope)
Given the upper bound on arrival time $\tau_d$, the envelope $Env$ of the query $(s_o,s_d,\tau_q)$ is the subset of connections $Env \subseteq C$ such that every connection
$(s_{dep}, \tau_{dep}, s_{arr}, \tau_{arr},t) \in Env$ satisfies:
(a) $sp_G(s_0,s_{dep}) + (\tau_{arr} - \tau_{dep}) + sp_G(s_{arr},s_d) \leq \tau_d - \tau_q$, 
(b) $\tau_{arr} + sp_G(s_{arr}, s_d) \leq \tau_d$, and
(c) $\tau_q \leq \tau_{dep}$.
\end{defn}

Condition (a) ensures the envelope includes connections that could potentially form a journey with travel time no longer than the current journey's travel time, $\tau_d - \tau_q$, regardless of delays. %
Condition (b) ensures the envelope includes connections that could potentially form a journey arriving at the destination no later than the current optimal journey.
Condition (c) restricts the envelope to connections that depart at or after the query time, considering any delays they might experience.
Constructing this envelope is straightforward. 
It involves running two Dijkstra searches on the $TIG$: a forward search from the origin stop $s_o$ and a backward search from the destination stop $s_d$, to compute $sp_G(s_o,s)$ and $sp_G(s,s_d)$, respectively, for every stop $s \in TIG$.
Using these shortest path durations, only connections in $C$ satisfying all conditions (a), (b), and (c) are added to the envelope.

\begin{example}
Consider the query $(s_1$,$s_6$,8:00) in Example~\ref{ex:motiv} for the network in Figure~\ref{fig:example}. The query's envelope includes the connections on the dashed route $r_1$, the solid route $r_2$ from $s_3$ to $s_6$ for the trip departing $s_2$ at 8:00, and the dotted route $r_3$ from $s_5$ to $s_6$ for the trip departing $s_8$ at 8:00. 
\end{example}

\ignore{Once the modified Dijkstra searches from both sides are completed, the algorithm extracts the envelope by including a vertex and its incoming and outgoing edges if: (i) the vertex has been visited by both modified Dijkstra searches from $s_o$ and $s_d$; and (ii) there exist tentative travel times $\tau_{s_o}$ and $\tau_{s_d}$ from the searches such that their sum is less than or equal to $\tau_{d} - \tau_q$ (i.e., $\tau_{s_o} + \tau_{s_d} \leq \tau_{up}$). The algorithm terminates and returns the envelope after checking each vertex.
}

\begin{theorem} \label{theo::correctness}
    There is no journey $j(s_o, s_d)$ from $s_o$ to $s_d$, starting at or after $\tau_q$ and using connections outside the envelope $Env$, that can arrive before $\tau_d$, regardless of delays.
    
    \begin{proof}
        Suppose, to the contrary, that given some delay events $EV$, there exists a journey $j = \langle c_0, \dots, c_n \rangle$ from $s_o$ that arrives at $s_d$ before $\tau_d$, where some connection $c_i$ is not in the envelope $Env$.
        Since $c_i = (s_{dep}, \tau_{dep}, s_{arr}, \tau_{arr},t)$ is not in $Env$, we have one of the following cases:
        (a) $sp_G(s_0,s_{dep}) + (\tau_{arr}- \tau_{dep}) + sp_G(s_{arr},s_d) >
        \tau_d - \tau_q$;
        (b) $\tau_{arr} + sp_G(s_{arr},s_d) > \tau_d$; or
        (c) $\tau_q >$ $\tau_{dep}$.
        In case (a), by definition, the sum of the durations of connections $c_0, \ldots, c_{i-1}$ is greater than or equal to $sp_G(s_0,s_{dep})$, and the sum of the durations of connections $c_{i+1}, \dots, c_n$ is greater then or equal to
        $sp_G(s_{arr},s_d)$. 
        Using connection $c_i$ then requires at least $\tau_{arr} - \tau_{dep}$ time regardless of delay $\delta$; hence, the arrival time at $s_d$ must be greater than $\tau_d$. Contradiction. 
        In case (b), by definition, a journey using connection $c_i$ cannot arrive at $s_{arr}$ earlier than $\tau_{arr}$, and then there is no journey from $s_{arr}$ to $s_d$ that takes less than $sp_G(s_{arr},s_d)$. Contradiction.
        For case (c), any connection $c_i$ cannot be reached if its departure time, adjusted for any delay $\delta$, is earlier than the current planning time $\tau_q$. Contradiction.
        Since all cases lead to a contradiction, the theorem holds.
    
    \end{proof}

\end{theorem}

It is worth noting that while our approach focuses on handling the common case of disruptions, where a delay in one connection propagates to subsequent connections on the same trip, it can also handle the less common case where a vehicle catches up and recovers from the delay in these connections.
The envelope remains valid as long as the sped-up connection does not arrive earlier than its scheduled time or complete in less time than its minimum scheduled duration.

\begin{algorithm2e}[t]
\small
\setcounter{AlgoLine}{0}
    \SetKwInput{Input}{Input}
    \SetKwInput{Output}{Output}
    \SetKwInput{Initialisation}{Initialisation}
    \Input{$s_{curr}$: current stop of user; $\tau_{curr}$: current time/departure time; $s_d$: destination stop; 
    $j_{curr}:$ current journey from $s_{curr}$ to $s_d$.
    $Env$: envelope of current journey;
    }
    \Output{
    $j_{curr}$: residual journey to $s_d$; 
    $Env$: updated envelope;
    }
    $Env = $ update the connections at current time $\tau_{curr}$; \label{algo2:line:updateEnv}\\ 
    \If{$s_{curr} = s_o$ or $j_{curr}$ is delayed}{ \label{algo2:if1}
        $(j_{curr}, Env)$ = 
        pull$(s_{curr},s_d,\tau_{curr})$; \% Call the server \label{algo2:pull}
    }
    \ElseIf{$Env$ is delayed}{ \label{algo2:if2}
        Sort the connections in $Env$ based on departure time; \\ \label{algo2:line:sortEnv}
        $j_{curr} = $ run CSA on $Env$ to plan $j(s_{curr}, s_d, \tau_{curr})$; \\
        \label{algo2:line:runCSAEnv}
    }
    \textbf{return} $(j_{curr}$, remove $c_0(j_{curr})$ from $j_{curr}$, $Env$); \label{algo2:return}
    \caption{Envelope Replanning Algorithm (Push)}
    \label{algo:env}
    \normalsize
\end{algorithm2e}

\subsubsection{Replanning on the Envelope}

Utilising the defined envelope, the edge device efficiently runs CSA within it to replan the next connection for the user to execute. 
Algorithm~\ref{algo:env} presents the pseudocode of our envelope replanning algorithm.
Given an initial query $(s_o,s_d,\tau_q)$, Algorithm~\ref{algo:env} is recursively called to determine the best journey to execute, starting from the origin stop $s_o$ until the destination stop $s_d$ is reached. 
During each invocation, the algorithm takes as input the current stop $s_{curr}$, the current time $\tau_{curr}$, the destination stop $s_d$, the current journey $j_{curr}$, and the envelope of the current journey, $Env$. 
Initially, $s_{curr}$ and $\tau_{curr}$ are set to $s_o$ and $\tau_q$, 
while $j_{curr}$ and $Env$ are empty and will be computed in the first iteration.

At each replanning step, the algorithm first updates connections in the envelope $Env$ to incorporate delays known at the current time $\tau_{curr}$, rather than updating all connections $C$ in the timetable (line~\ref{algo2:line:updateEnv}). 
Based on the this update, there are three possible scenarios for the next move:

\begin{enumerate}
    \item The current journey $j_{curr}$ is delayed, or the algorithm is being called for the first time (i.e., $s_{curr} = s_o$). 
    $j_{curr}$ is considered delayed if any of its transfers are missed or if its final leg\footnote{the connections from the final transfer stop to $s_d$.} is delayed.
    In this case, the current envelope $Env$ is no longer valid because the arrival time $\tau_{d}$ of the optimal journey is pushed back. 
    Algorithm~\ref{algo:env} replans the optimal journey for $(s_{curr}, s_d, \tau_{curr})$ by calling the server, 
    rebuilding the envelope accordingly (line~\ref{algo2:pull}).

    \item The current journey $j_{curr}$ is not delayed, but delays exist in other connections within the envelope $Env$.
    In this case, $Env$ remains valid. The algorithm resorts the connections within $Env$ and runs CSA on $Env$ to plan an optimal journey for $(s_{curr}, s_d, \tau_{curr})$ (lines~\ref{algo2:line:sortEnv} – \ref{algo2:line:runCSAEnv}).

    \item Neither the current journey $j_{curr}$ nor the envelope $Env$ is delayed. 
    In this case, $j_{curr}$ preserves optimality, and no replanning is needed.
\end{enumerate}

Case (1) is the most time-consuming part of the algorithm, requiring a server call to create a new journey and envelope (Algorithm~\ref{algo:naive}).
The function buildEnvelope$(s_{curr},s_d,\tau_{curr},\tau_d(j))$ constructs the envelope for further replanning using the new arrival time $\tau_d(j)$ (line~\ref{algo1:env}, Algorithm~\ref{algo:naive}). 
Fortunately, as shown in the experiments, case (1) occurs infrequently.
The envelope is constructed during the first algorithm call and may need to be reconstructed once or twice in most cases.

Finally, after replanning $j_{curr}$ according to the different cases, the algorithm returns $j_{curr}$ including its first connection, $c_0$, as the next action for the user to execute (line \ref{algo2:return}).
Additionally, the algorithm provides the remaining part of $j_{curr}$ (excluding $c_0$) and the updated envelope $Env$ for the next replanning iteration.
To prepare for this iteration, the algorithm sets $\tau_{curr}$ to $\tau_{arr}(c_0)$ and $s_{curr}$ to $s_{arr}(c_0)$.


\section{Experiments}

\subsection{Experiments Setup}
All experiments were implemented in C++17 with full optimisation on a 3.20 GHz Apple M1 machine with 16 GB of RAM, running macOS 14.5 and utilising a single thread.

\paragraph{Datasets.}
Four metropolitan networks of varying sizes, namely Perth, Berlin, Paris, and London, are considered.
Each dataset is based on a weekday timetable and includes all available public transport modes, such as trains, subways, trams, buses, and ferries. 
All datasets were imported in the General Transit Feed Specification (GTFS) format.
The first two datasets are from an open data platform,\footnote{https://openmobilitydata.org}
while the others are sourced from \cite{phan2019fast}.\footnote{https://files.inria.fr/gang/graphs/public\_transport} 
Table \ref{table:datasets} summarises the key metrics for the datasets.
To model transfers, the original footpath set from the source is used.
Additional footpaths are created to ensure the footpath graph is transitively closed. 
Loop footpaths are also added for stops that do not have any. 
Numbers in Table \ref{table:datasets} include these footpaths.

\begin{table} [t]
    \renewcommand{\arraystretch}{0.75}
    \setlength\tabcolsep{7pt}
    \centering
    \small
    \caption{Key metrics for the test datasets}
    \begin{tabular} {l|r|r|r|r}
        \toprule
        \textbf{Dataset} & \textbf{Stops} & \textbf{Connections} & \multicolumn{1}{c|}{\textbf{Trips}} & \textbf{Footpaths} \\
        \midrule
        Perth & 14,022 & 643,737 & 21,130 & 17,689 \\
        Berlin & 27,941 & 1,449,971 & 70,941 & 76,456 \\
        Paris & 19,800 & 1,643,608 & 71,407 & 73,850 \\
        London & 19,746 & 4,572,979 & 121,760 & 46,566 \\
        \bottomrule
    \end{tabular}
    \label{table:datasets}
\end{table}

\paragraph{Delay Modelling.}
Delay events are modelled as described in Section \ref{sec:Pre}. 
A real delay probability distribution could not be applied due to the lack of historical delay data. 
As a result, we use a synthetic delay model that follows an exponential distribution, as suggested by some studies \cite{hansen2001improving,markovic2015analyzing}.
This distribution has a single parameter $\lambda = 1 / \bar{\delta}$, where $\bar{\delta}$ is the mean delay.
The Cumulative Distribution Function (CDF) is given by $P(\delta \leq x) = 1 - e^{-\lambda x}$, where $\delta$ is the duration of delay experienced by a trip $t$, and $x \geq 0$.
Very short delays below 30 seconds are ignored.
The mean delay values used vary depending on transport mode and time of day, and were estimated based on performance reports for London.\footnote{https://tfl.gov.uk/corporate/publications-and-reports}
For fully-separated modes, such as trains and subways, $\bar{\delta} = 2$ minutes is used regardless of the time of day.
For semi-separated modes, such as trams, $\bar{\delta} = 3$ minutes is used for off-peak periods and $\bar{\delta} = 7$ minutes for peak periods.
Finally, for mixed-traffic modes, such as buses, $\bar{\delta} = 5$ minutes is used for off-peak periods and $\bar{\delta} = 10$ minutes for peak periods.
The peak and off-peak periods are determined based on the distribution of connections throughout the day.
The exact periods are highlighted in the experimental figures.

\paragraph{Query Generation.}
We generate a sample of 1,000 unique stop pairs ($s_o$, $s_d$) chosen uniformly at random for each dataset.
These stop pairs are assigned ten fixed departure times ($\tau_q$) throughout the day, from midnight to 9 pm.
This results in a total of 10,000 queries per dataset.
This sample size enables realistic estimates of average query runtime and time savings distribution, and aligns with common practice in the field.
Note that the stop pairs were verified to ensure a dynamic replanning solution exists for all ten departure times.

\begin{figure*}[t]
\centering
\scriptsize
\begin{tabular}{@{}c@{}c@{}c@{}c@{}}

\begin{minipage}{.25\textwidth}
\includegraphics[width=\textwidth]{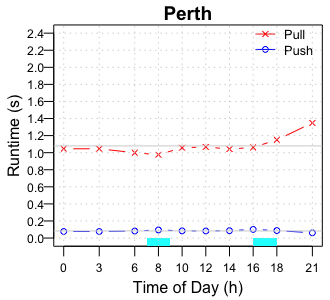}
\end{minipage}&
\begin{minipage}{.25\textwidth}
\includegraphics[width=\textwidth]{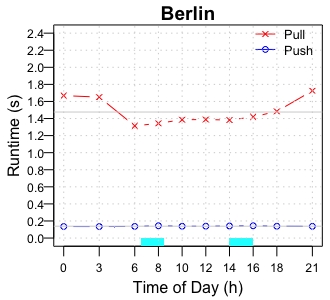}
\end{minipage}&
\begin{minipage}{.25\textwidth}
\includegraphics[width=\textwidth]{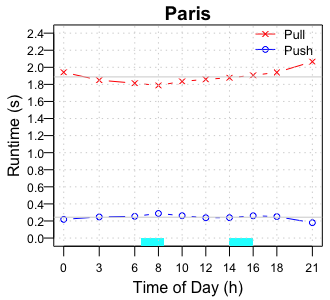}
\end{minipage} &
\begin{minipage}{.25\textwidth}
\includegraphics[width=\textwidth]{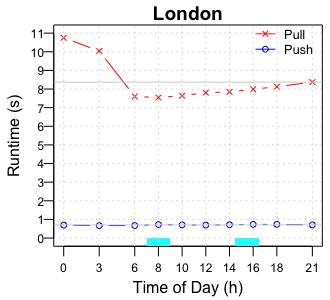}
\end{minipage}

\end{tabular}
\caption{
Runtime comparison between the pull and push approaches throughout the day. 
Different y-axis limits are used to enhance data clarity across plots.
Cyan blocks represent peak periods, while grey lines indicate overall average runtimes.
}
\label{fig:runtime}
\end{figure*}

\begin{figure*}[t]
\centering
\scriptsize
\begin{tabular}{@{}c@{}c@{}c@{}c@{}}

\begin{minipage}{.25\textwidth}
\includegraphics[width=\textwidth]{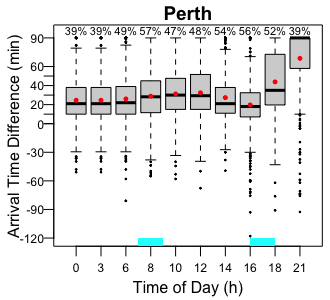}
\end{minipage}&
\begin{minipage}{.25\textwidth}
\includegraphics[width=\textwidth]{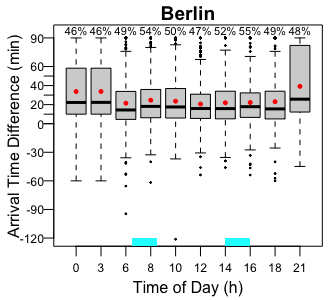}
\end{minipage}&
\begin{minipage}{.25\textwidth}
\includegraphics[width=\textwidth]{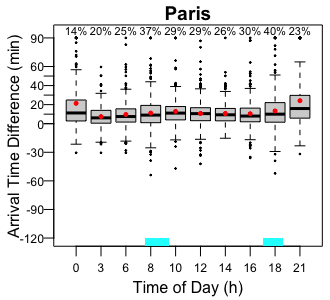}
\end{minipage} &
\begin{minipage}{.25\textwidth}
\includegraphics[width=\textwidth]{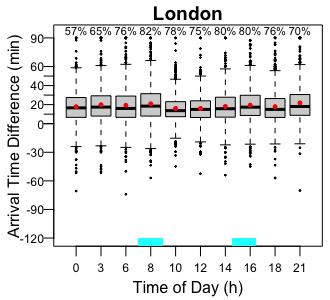}
\end{minipage}

\end{tabular}
\caption{
Distribution of difference in arrival time at destination for affected queries (percentages shown) between static planning (SP) and dynamic replanning (DR) across the day. 
Red dots indicate mean values, while cyan blocks indicate peak periods.}
\label{fig:Delays}
\end{figure*}

\subsection{Experiment 1: Query Performance}

The mean query runtimes for both the baseline pull approach and the proposed push approach are presented in Figure \ref{fig:runtime}.
This runtime reflects the average total time required to process a query from the source to the target, including replanning at each intermediate stop along the journey---whether handled solely by the server (pull approach) or shared with the edge device (push approach).
The results demonstrate the efficiency of the push approach, achieving runtimes well within a fraction of a second, while the pull approach requires several seconds.
The push approach consistently achieves an order-of-magnitude speedup across all datasets and times of day.
This translates to the push approach being capable of handling 10 times more queries than the pull approach within the same time period.
The relatively slower runtimes for the London dataset are due to its larger number of connections.
The pull approach performs worse overnight due to the low frequency of relevant services, especially when a transfer is missed, leading to scanning more unnecessary connections.

Table~\ref{table:stats} shows statistics for the push approach across datasets, including the average envelope size and the average percentage of stops per query where delays affect the current journey, the envelope, or neither.
Two key factors contribute to the superior performance of the push approach:
(i) the significantly smaller size of the envelope compared to the full connections array (around 5\%), which considerably reduces the search space, and
(ii) the infrequent need for server calls to reconstruct the envelope (about 5\% of the intermediate stops).
This low frequency means the server receives approximately 20 times fewer calls compared to the pull approach (where a request is made at every stop), enabling a significantly higher query handling capacity.
Finally, the statistics indicate that the amount of data sent from the server to each edge device, which represents the total envelope size and its updates throughout the journey, is modest, averaging only about 3~MB.
This demonstrates the practicality of the proposed push approach for real-world applications.

\subsection{Experiment 2: Arrival Time Saving}
To examine travel time savings from dynamic replanning (DR), we develop three alternative baseline strategies for handling disruptions.
First, \emph{static planning} (SP) represents the default strategy, where users plan their journey using the published delay-free timetable. 
They follow the plan as closely as possible despite any delays.
If the plan fails due to a missed transfer, users attempt to repair the static plan with minimal deviation. 
They wait at the same stop and take the next available train to the next transfer stop.
Second, in the \emph{Snapshot Replanning} (SR) strategy, users (re)plan their journey at $\tau_q$ based on the most recent delay updates just before starting from the origin stop, and then react to delays as in static planning. 
Finally, in the \emph{journey-delayed replanning} (JDR) strategy, replanning occurs only when the current journey is delayed; essentially removing lines \ref{algo2:if2}-\ref{algo2:line:runCSAEnv} from Algorithm~\ref{algo:env}. 
JDR can be understood as an improvement on contingent planning, incorporating real-time delays and replanning dynamically.

\paragraph{DR vs. SP.} Figure~\ref{fig:Delays} illustrates the difference in arrival time at the destination between dynamic replanning and static planning throughout the day, excluding queries with the same arrival time in both scenarios. 
For queries that arrive extremely late in SP due to a lack of available services and require waiting until the next day, we assume users will not wait indefinitely before taking alternative actions (e.g., taking a taxi).
In such cases, users incur a penalty of 90 minutes for inconvenience and extra time and cost, which means they are considered to arrive 90 minutes later than with DR.
The results show that a significant portion of journeys are affected, with almost half in both Perth and Berlin, slightly fewer in Paris, and more in London. 
The mean difference in arrival time for affected journeys ranges from 20 to 30 minutes of time savings in both Perth and Berlin, and from 5 to 20 minutes in both Paris and London.
Note that in a few cases, around 4-5\% of journeys, DR may result in a later arrival than SP.
This occurs when the updated plan advised by DR encounters significant unforeseen delays later.
The findings also show that affected journeys during overnight hours (i.e., 0, 3, and 21) are generally fewer but result in larger time savings compared to the rest of the day. 
This is because services are less frequent during these hours, so even if a journey is delayed, transfers are more likely to remain valid.
However, if a transfer is missed, the consequences can be significant, leading to longer waits and late arrivals.

\paragraph{DR vs. SR and JDR.} Table~\ref{table:savings} shows the advantages of dynamic replanning compared to both the snapshot replanning and journey-delayed replanning scenarios.
SR can provide a limited improvement over SP. 
Considering JDR, a significant part of the time savings of DR comes from replanning even when the current journey is not even delayed.
This demonstrates the importance of the dynamic replanning in minimising the impact of disruptions.

\begin{table} [t]
    \renewcommand{\arraystretch}{0.75}
    \setlength\tabcolsep{7pt}
    \centering
    \small
    \caption{Envelope statistics: mean proportion of connections, mean amount of pushed data, and percentage of stops per query across all delay cases.}
    \begin{tabular} {l|r|r|r|r}
        \toprule
        \multicolumn{1}{c|}{\textbf{Metric}} & \textbf{Perth} & \textbf{Berlin} & \textbf{Paris} & \textbf{London}\\
        \midrule
        Envelope Size & 5.9\% & 3.7\% & 7.5\% & 2.8\% \\
        \midrule
        Pushed Data (MB) & 1.4 & 1.6 & 4.1 & 4.8 \\
        \midrule
        Journey Delayed & 3.7\% & 4.8\% & 4.2\% & 6.0\% \\
        Envelope Delayed & 73.6\% & 81.0\% & 94.0\% & 87.5\% \\
        Neither Delayed & 22.7\% & 14.2\% & 1.8\% & 6.5\% \\
        \bottomrule
    \end{tabular}
    \label{table:stats}
\end{table}

\begin{table} [t]
    \renewcommand{\arraystretch}{0.75}
    \setlength\tabcolsep{4.3pt}
    \centering
    \small
    \caption{
    Average arrival time savings (Avg. Saving) in minutes from DR for affected queries (Q. Affected), compared to SR and JDR.}
    \begin{tabular} {c|l|c|c|c|c}
        \toprule
        \textbf{Comparison} & \multicolumn{1}{c|}{\textbf{Metric}} & \textbf{Perth} & \textbf{Berlin} & \textbf{Paris} & \textbf{London}\\
        \midrule
        \multirow{2}{*}{DR vs. SR} & Q. Affected & 42.3\% & 41.1\% & 20.4\% & 66.2\% \\
        & Avg. Saving & 30.2 & 24.2 & 10.6 & 16.0 \\
        \midrule
        \multirow{2}{*}{DR vs. JDR} & Q. Affected & 10.0\% & 13.1\% & 13.3\% & 21.1\% \\
        & Avg. Saving & 5.7 & 4.6 & 6.0 & 4.9 \\
        \bottomrule
    \end{tabular}
    \label{table:savings}
\end{table}


\section{Conclusion}

We formulated dynamic replanning in public transport routing to address disruptions.
A simple pull approach was proposed, followed by an advanced push approach that efficiently utilises resources via local replanning and focuses on relevant timetable parts with envelope construction.
Experiments showed the query efficiency of the push approach, and the effectiveness of dynamic replanning in saving travel time.

\bibliographystyle{named}
\bibliography{ijcai25}

\end{document}